# Public Acceptance of Cybernetic Avatars in the service sector:

# Evidence from a Large-Scale Survey in Dubai

Laura Aymerich-Franch[1], Tarek Taha[1], Takahiro Miyashita[2], Hiroko Kamide[3], Hiroshi Ishiguro[4], and Paolo Dario[1,5]

Affiliations:

[1] Dubai Future Labs, Dubai Future Foundation, Dubai, United Arab Emirates

[2] ATR (Advanced Telecommunications Research Institute International), Interaction Technology Bank, Keihanna Science City, Kyoto, Japan

[3] Graduate School of Law, Kyoto University,Kyoto, Japan

[4] Department of Systems Innovation, Osaka University, Osaka, Japan and ATR Hiroshi Ishiguro Laboratories

[5] The BioRobotics Institute, Scuola Superiore Sant'Anna, Pisa, Italy

Corresponding author:

Laura Aymerich-Franch

Dubai Future Labs, Dubai Future Foundation, Emirates Towers - Office Towers (Ground Level), Dubai, UAE

Laura.aymerich@dubaifuture.gov.ae

Declaration of Interest statement

The authors declare no competing interests

Acknowledgments

Research supported by Dubai Future Foundation and JST Moonshot R&D Grant Number JP- MJMS2011

# Public Acceptance of Cybernetic Avatars in the service sector:

# Evidence from a Large-Scale Survey in Dubai

**Abstract**

Cybernetic avatars are hybrid interaction robots or digital representations that combine autonomous capabilities with teleoperated control. This study investigates the acceptance of cybernetic avatars in the highly multicultural society of Dubai, with particular emphasis on robotic avatars for customer service. Specifically, we explore how acceptance varies as a function of robot appearance (e.g., android, robotic-looking, cartoonish), deployment settings (e.g., shopping malls, hotels, hospitals), and functional tasks (e.g., providing information, patrolling). To this end, we conducted a large-scale survey with over 1,000 participants. Overall, cybernetic avatars received a high level of acceptance, with physical robot avatars receiving higher acceptance than digital avatars. In terms of appearance, robot avatars with a highly anthropomorphic robotic appearance were the most accepted, followed by cartoonish designs and androids. Animal-like appearances received the lowest level of acceptance. Among the tasks, providing information and guidance was rated as the most valued. Shopping malls, airports, public transport stations, and museums were the settings with the highest acceptance, whereas healthcare-related spaces received lower levels of support. An analysis by community cluster revealed among others that Emirati respondents showed significantly greater acceptance of android appearances compared to the overall sample, while participants from the 'Other Asia' cluster were significantly more accepting of cartoonish appearances. Our study underscores the importance of incorporating citizen feedback into the design and deployment of cybernetic avatars from the early stages to enhance acceptance of this technology in society.

**Keywords:** cybernetic avatars, robot avatars, technology acceptance, multicultural contexts, social robots, human-robot interaction

# 1. Introduction

Cybernetic avatars are hybrid interaction robots or digital representations that can perform tasks on their own, while also being controlled by a human operator (Horikawa et al., 2023; Ishiguro, 2021; Ishiguro et al., 2025). Operators can interact socially through cybernetic avatars, control the avatar body, and communicate verbally through it (Aymerich-Franch et al., 2020; Dafarra et al., 2024; Hatada et al., 2024; Kishore et al., 2016). These entities are envisioned as a technology that can free humans from limitations of body, brain, space, and time (Ishiguro, 2021; Ishiguro et al., 2025). Robot avatars are primarily explored in real-world applications as surrogate bodies for individuals with reduced mobility (Hatada et al., 2024) and for remote travel and physical interaction, such as virtual museum visits (Roussou et al.,

2001). Additionally, virtual avatars are extensively used for real-world applications in gaming and interaction in social virtual reality applications, improving mental health, and promoting behavioral change (Aymerich-Franch, 2020a; Aymerich-Franch et al., 2014; Falconer et al., 2016; Rosenberg et al., 2013; Slater et al., 2019; Yee & Bailenson, 2007).

Our study examines social acceptance of cybernetic avatars, with a particular emphasis on robotic avatars for customer service. Acceptance is defined as the cybernetic avatar being willingly incorporated into the society (Broadbent et al., 2009). Cybernetic avatars have been progressively deployed in Japan; however, acceptance of this technology in other demographic contexts might differ (Kamide et al., 2025). Dubai presents a particularly valuable case study to examine acceptance of cybernetic avatars in multicultural societies, as over 200 nationalities live and work in the UAE (UAE Ministry of Foreign Affairs, 2024). In fact, the Emirati population of Dubai is estimated to be about 8%, and 92% are non-Emirati (Dubai Statistics Center, 2024).

Dubai's demographically diverse environment provides a rare opportunity to examine the acceptance of cybernetic avatars across highly varied geographical backgrounds within a single urban setting. This study capitalizes on the Emirate's cosmopolitan character to present a uniquely global perspective on the social acceptance of this emerging technology.

To gain a comprehensive understanding of cybernetic avatar acceptance we conducted a large-scale survey involving over 1,000 participants. To our knowledge, it constitutes the largest study to date investigating the acceptance of cybernetic avatars within such a culturally heterogeneous population.

## 1.1. Avatar appearance and modality

At present, cybernetic avatars present two primary modalities: physical robot avatars and virtual reality avatars (Aymerich-Franch, 2020b; Horikawa et al., 2023; Ishiguro et al., 2025). The first research question explored acceptance of the two modalities:

RQ1. To what extent are the two main modalities of cybernetic avatars (robot and virtual) accepted, among Dubai residents?

To address RQ1, survey participants were asked to envision an ideal society for Dubai and indicate on a three-point scale (disagree – neutral – agree) whether avatars designed to assist customers in the service sector would take the form of physical robots and/or virtual representations.

The appearance of robots has been extensively discussed in the literature as a key factor influencing their acceptance (Hameed et al., 2016). The second research question explored acceptance in relation to different robot avatar appearances:

RQ2. To what extent are different robot avatar appearances accepted among Dubai residents?

To explore RQ2, we developed a typology of six distinct categories that represent the primary social robot appearances currently being used in the service sector worldwide based on previous works (Aymerich-Franch & Ferrer, 2020, 2023, Glas et al., 2016; Nakata et al., 2022). The final typology included six groups: ultra-realistic android, hybrid android (human-robotic looking), highly anthropomorphic robotic looking, low anthropomorphic robotic looking, cartoonish looking, and animal looking. We then selected two visual representations which exemplified each category in the survey. For the two android categories, we borrowed images from Erica and Ibuki, which are androids designed by Ishiguro and colleagues (Glas et al., 2016; Nakata et al., 2022). For the remaining categories, we generated ad hoc designs with the aid of graphic designers and AI tools. Figure 1 shows the robot avatar pictures and designs that exemplified each category in the survey. Participants were asked to envision an ideal society for Dubai and indicate whether robot avatars designed to assist customers in the service sector would have each of the appearances.

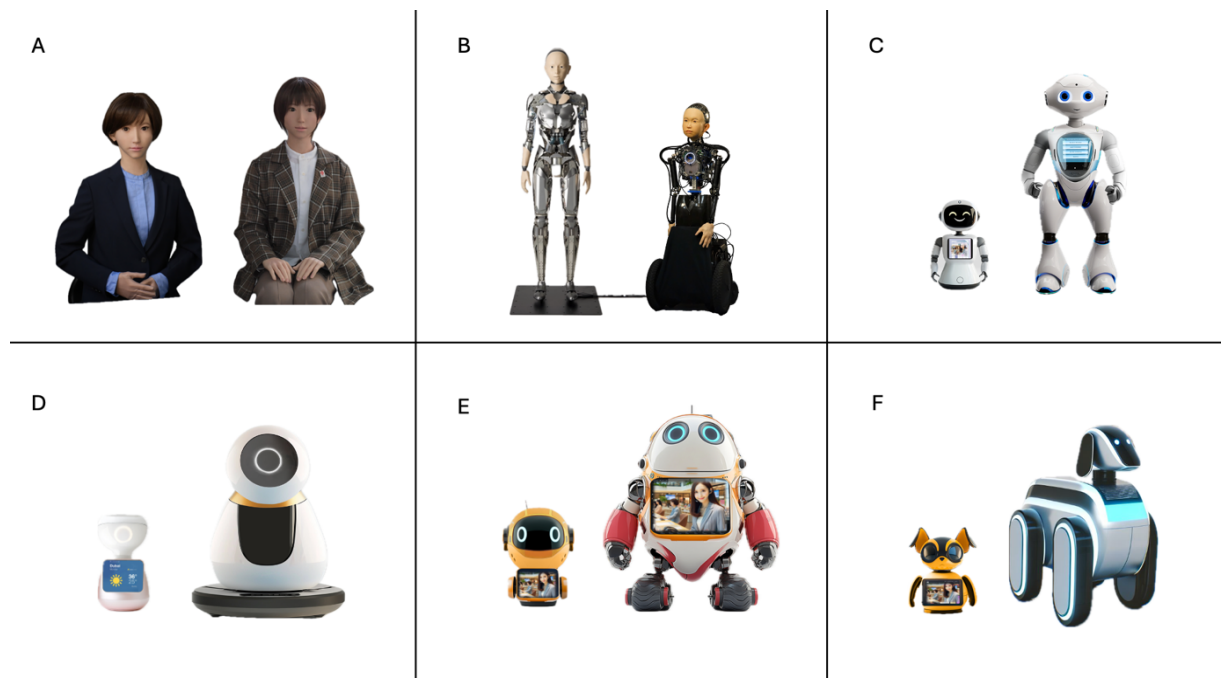

*Figure 1.* Robot avatar appearances assessed in the survey

## 1.2. Spaces in the service sector

The third research question explored robot avatar acceptance depending on the setting where robot avatars are found:

RQ3. To what extent are robot avatars accepted in the different spaces of the service sector, among Dubai residents?

We elaborated a list of key spaces in healthcare, education, financial, retail, transportation, hospitality, and government services where robot avatars could potentially be introduced, based on the environments where social robots for customer service are commonly deployed in real-world scenarios, as identified by Aymerich-Franch and Ferrer (2020, 2022, 2023). Participants were asked whether, in the ideal future society they envisioned for Dubai, robot avatars would be permitted in these spaces (e.g., shopping malls, hospitals, schools).

## 1.3. Tasks for the service sector

We also explored robot avatar acceptance in Dubai depending on the task they perform:

RQ4. To what extent are robot avatars accepted to perform the different tasks typically performed by social robots, among Dubai residents?

For that, we elaborated a list of general tasks, drawing on the roles that social robots currently fulfill in real-world settings, as identified by Aymerich-Franch and Ferrer (2020, 2022, 2023). The functions included providing information to customers, customer registration / check in – out, providing indications and guidance to find a place, telepresence to speak with the human controlling the robot, patrolling spaces for security, companionship and entertainment functions, object delivery, on-site customer data collection and feedback collection. We asked participants whether in the ideal future society they envisioned for Dubai, robot avatars would be used for these tasks in the service sector.

## 1.4. Relationship with robots and demographic variables

We additionally collected information related to relationship with robots. In particular, we included general level of interest in scientific discoveries and technological developments, general view of robots and attitudes towards robots scale (items based on Aymerich-Franch and Gómez (2024) and European Commission (2012), and fear of robots scale (items based on Aymerich-Franch and Gómez, 2024 and Liang and Lee, 2017).

Regarding demographics, we collected information on nationality, gender, age, occupation, time living in the UAE, born in the UAE, level of studies, income level, religious beliefs, background in Computer Science, level of programming skills, experience using robots, experience interacting with social robots and robot avatars, and collectivism – individualism orientation (Triandis & Gelfand, 1998).

## 2. Methodology

### 2.1. Contents of the survey

The survey contained the following blocks[*]:

- Introduction to cybernetic avatars
- Acceptance of robot avatars by appearance (ultra-realistic androids, hybrid androids, highly anthropomorphic robotic-looking, low anthropomorphic robotic looking, cartoonish looking, and animal looking)
- Acceptance of robot avatars by space (e.g., shopping malls, banks, etc.)
- Acceptance of robot avatars by task (e.g., provide information, patrolling, etc.)
- Acceptance of cybernetic avatars by modality (robotic and virtual)
- Relationship with robots and technology (interest in science and technology, attitudes towards robots, fear of robots)
- Demographics

We additionally added two attention check questions among the previous questions (e.g., "I am paying attention to the survey, select "XX").

### 2.2. Platform

The survey was outsourced to an external market research firm, which was responsible for recruiting participants and administering the survey through its platform. The survey was conducted in English. Participants were compensated with points, which could be accumulated and exchanged for rewards. Data collection took place between December 2024 and February 2025. Upon completion of the survey, the market research firm provided the raw data in Excel format, which we used for subsequent analysis.

---

[*] During their participation, the participants completed additional blocks of questions related to acceptance of robot avatars for capability enhancement and the design of an ideal robot avatar for customer service which are not reported here.

## 2.3. Participants

The survey was administered in English and it was open to participants aged 18 or older, residents of Dubai, from any of the 6 community clusters listed in Table 1. The survey was initiated by 4,000 participants. Of these, 1,542 were screened out due to ineligibility, 883 were excluded after quota targets were reached, 198 did not finalize the survey, and 376 were removed through quality control checks. These measures helped ensure a high-quality and demographically balanced dataset for analysis. The final sample consisted of 1001 participants. The study received ethical approval from the [institution hidden for peer review] Ethics Committee.

### 2.3.1. Community Clusters Distribution

Dubai presents an atypical demographic structure. The population size of the Emirate of Dubai is estimated at 3.8 million, having the highest population among the emirates in the UAE. Of them, 31.4% are females and 68.6% are males. The majority of the residents (58.49%) are aged between 25 to 44 years.

Over 200 nationalities live and work in the UAE (UAE Ministry of Foreign Affairs, 2024). Given that the objective of the study was to capture perspectives from all major communities residing in Dubai, rather than to obtain a statistically representative sample of the general population, we employed a structured sampling strategy. Specifically, a stratified sampling method was used, based on gender and community affiliation, to ensure balanced representation.

For that, we identified the expat communities with 10,000 residents or more in the UAE (Wikipedia, 2024). A total of 35 countries and areas were identified. We then classified the countries by geographical region following the classification provided by the Statistics Division of the United Nations for statistical use (UNSD, 2024) to get an understanding of the geographical distribution of the represented groups. The initial classification resulted in countries and areas in the following regions: Eastern, South-Eastern, Southern, and Western Asia; Northern and Sub-Saharan Africa; Eastern, Northern, Southern and Western Europe; Northern America; Australia and NZ. We then further organized the countries and regions into six main community clusters that best reflect Dubai's social composition: Emiratis (UAE citizens), a Middle East cluster (neighboring countries), a South Asian cluster (representing the largest expatriate group, comprising nearly 60% of the UAE population), and three additional clusters for 'Other Asian', 'Other African', and 'Western'.

## 3. Results

The final sample consisted of 1001 valid participants, 503 females and 498 males. Participants were distributed across six community clusters, each comprising between 157 and 174 individuals, as shown in Table 1. Of the participants, 4.1% were between 18–24 years old,

38.0% were 25–34, 38.4% were 35–44, and 19.6% were 45 or older. Most respondents (42%) had lived in Dubai between 4–10 years, with 24.1% born in the UAE. Over half (53.6%) indicated a high interest in scientific discoveries and technological developments, and the majority expressed a positive general view of robots, with 53.7% fairly positive, and 31.3% very positive. Fear of robots was moderate overall: 30.9% reported no fear, while 53.2% reported slightly or moderate fear. Most participants had occasionally interacted with robots (58.2%), and 54.1% had occasionally interacted with avatars.

| Cluster | Countries / Areas | Participants (N) |
|---|---|---|
| Emirati | United Arab Emirates | 170 (83 males, 87 females) |
| Middle East | Egypt, Iraq, Jordan, Lebanon, Palestine, Syria, Saudi Arabia, Türkiye, Qatar, Kuwait, Bahrain, Oman | 166 (84 males, 82 females) |
| Southern Asia | India, Pakistan, Bangladesh, Nepal, Sri Lanka | 173 (89 males, 84 females) |
| Other Asia | Philippines, China, Indonesia, Japan, Taiwan, Malaysia, Myanmar, Singapore, Kazakhstan, Uzbekistan, Armenia, Azerbaijan | 174 (81 males, 93 females) |
| Western | European Union (Austria, Bulgaria, Croatia, Cyprus, Finland, France, Germany, Greece, Ireland, Italy, Lithuania, Poland, Portugal, Romania, Sweden), United Kingdom and overseas territories, United States and territories, Russia, Canada, Australia, Switzerland, Belarus, Bosnia Herz, Ukraine, Serbia, New Zealand, Albania, Peru, Cuba, Panama, Dominican Rep. | 161 (85 males, 76 females) |
| Other Africa | Ethiopia, Kenya, Benin, Cameroon, Ghana, Central Africa Rep, Congo, Cote Ivoire, Eritrea, Gambia, Ghana, Guinea, Nigeria, Rwanda, Sierra Leone, Tanzania, Togo, Uganda, Zimbabwe, Comoros | 157 (76 males, 81 females) |

**Table 1.** *Clusters by country / areas, and sample representation in the large-scale survey.*

### 3.1. Cybernetic avatar acceptance by modality: digital vs robotic

Overall, acceptance of cybernetic avatars was high. Acceptance of physical robots was higher than that of digital formats such as virtual avatars presented on screens or in VR environments (Figure 2). Specifically, 67.3% of respondents agreed with the deployment of robots in customer-facing service roles, compared to 56.9% who agreed with the use of digital avatars. A Wilcoxon Signed-Rank Test showed a significant preference for robots over digital formats (Z = −8.91, p < .001).

We then examined acceptance by nationality groups (community clusters). Acceptance of physical robots was consistently high across clusters, with 74.1% of Emirati respondents agreeing, followed by 70.5% acceptance from the Middle East cluster, 67.1% acceptance from the South Asia cluster, 65.0% acceptance from the Other Africa cluster, 63.8% acceptance from Other Asian countries, and 63.4% acceptance from the Western cluster. Acceptance of digital avatars was slightly lower. In particular, agreement was highest among Emirati (64.7%) and Other Africa cluster (62.4%) participants, followed by Westerners (60.9%), Other Asians (55.2%), Middle Easterners (51.2%), and South Asians (48.0%).

To examine differences in agreement with digital and physical avatar formats across community clusters, we conducted binomial tests comparing each cluster's agreement rate to the overall sample proportion. Responses were recoded as "agree" vs. "neutral/disagree," and each cluster was tested separately for each format (six comparisons per format). No differences remained statistically significant after Bonferroni correction ($\alpha$ = .0083). Full results are reported in Table 1A & 2A (SI).

A Mann-Whitney U test was conducted to assess gender differences in the acceptance of robot avatars. Although the median score was the same for men and women (Mdn = 3), the distribution of responses differed significantly (U = 113,829.00, z = -3.03, p = .002), with men showing greater overall acceptance. No significant gender differences (Mdn = 3) were observed for digital avatars (U = 123,512.00, z = -0.43, p = .667).

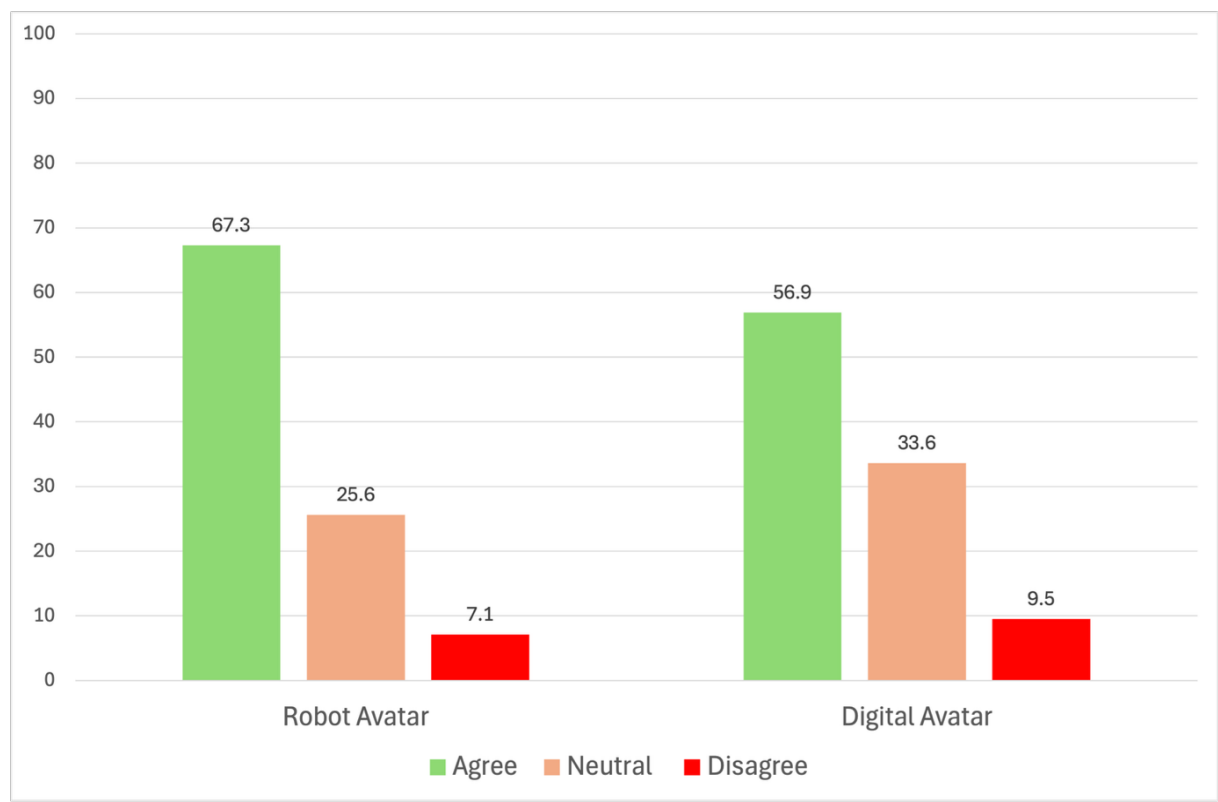

***Figure 2.*** *Acceptance rate by avatar modality: robotic and digital*

## 3.2. Cybernetic avatar acceptance by appearance

Overall, acceptance of robot avatars for customer service was very high, with the majority of respondents expressing agreement or neutrality across all appearance types. Robotic-looking avatars with high anthropomorphism received the highest level of agreement (61.1% agreed). This was followed by cartoonish (53.3%) and android designs (50.4%), both of which also received majority support. In contrast, robotic-looking avatars with low anthropomorphism (42.6%) and hybrid androids (41.2%) received more moderate support, potentially reflecting ambivalence toward designs that fall between clearly human or clearly robotic. Animal-looking robots were the least favored, with only 39% agreeing with their use in service roles, and the highest level of disagreement among all categories (28.0%), suggesting lower acceptability for zoomorphic forms in this context. Overall, the data suggest that the public in Dubai tends to prefer robot appearances that are distinctly anthropomorphic either in a robotic, human-like or cartoonish form, with less acceptance for ambiguous or zoomorphic designs (Figure 3).

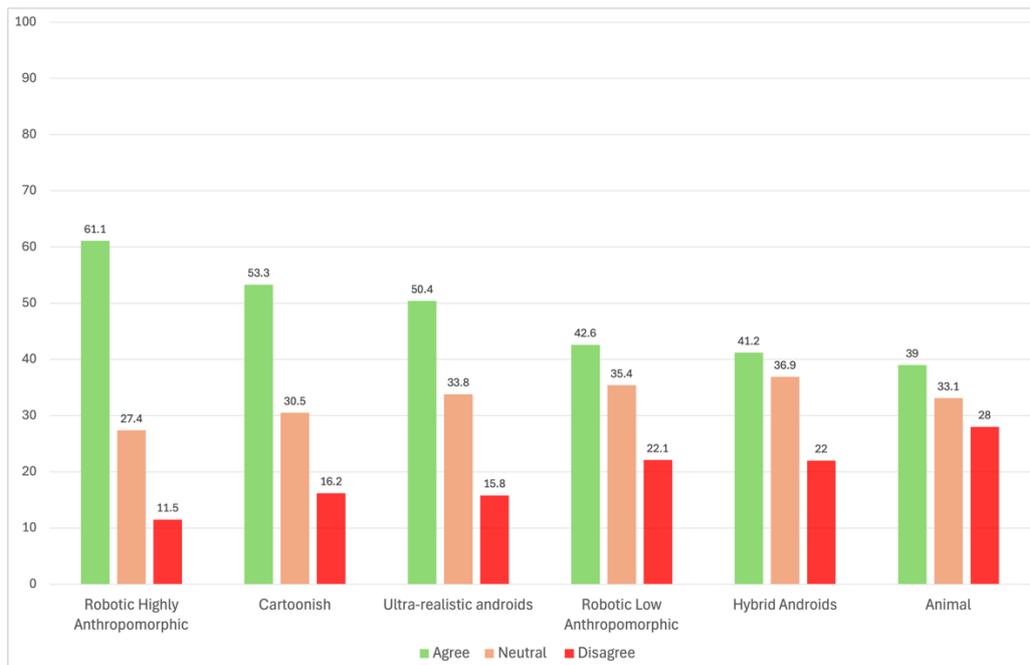

*Figure 3.* Acceptance rate by Appearance

### 3.2.1. Appearance by community and gender

To evaluate whether the acceptance of the different robot avatar appearances differed significantly across community clusters, we conducted binomial tests comparing the proportion of "agree" responses within each cluster to the corresponding agreement rate in the overall sample (50.4%). Responses were recoded into a binary outcome ("agree" = 0 vs.

"not agree" = 1, combining "neutral" and "disagree") for consistency with test assumptions. The tests were performed independently for each of the six clusters, and a Bonferroni correction was applied to control for multiple comparisons, setting the adjusted alpha level at .0083. "Agree" responses were treated as the successes in the test.

Results showed that the Emirati cluster exhibited a significantly higher agreement rate with the Android appearance (69.4%) than the overall sample, $p < .001$. Conversely, the Other Asian cluster demonstrated a significantly lower agreement rate (33.3%) compared to the baseline, $p < .001$. No other clusters differed significantly from the overall proportion under the corrected threshold (Table 3A – SI). For the hybrid android appearance, the Other Asian cluster reported a significantly lower agreement rate (31.6%) compared to the overall sample, $p = .006$, which remained significant after correction (Table 4A – SI). No other cluster differed significantly from the baseline. For the Cartoonish appearance (Table 5A – SI), the Other Asian cluster showed significantly higher acceptance (68.4%) compared to the overall sample, which remained significant after correction ($p < .001$). Additionally, the Western cluster showed significantly lower acceptance (40.4%), which was also significant after correction ($p < .001$). Results for the Robotic High Anthropomorphic, Robotic Low Anthropomorphic, and Animal-like appearance formats showed no statistically significant deviations from the overall acceptance rates in any cluster (Table 6A, 7A, 8A - SI).

To assess whether acceptance of different robot avatar appearance types varied by gender, we conducted Mann–Whitney U tests comparing male and female participants' responses across the six appearance types (Table 9A – SI). Each response was treated as ordinal (Disagree = 1, Neutral = 2, Agree = 3). A Bonferroni correction was applied to account for multiple comparisons across the six appearance types (adjusted $\alpha = .0083$). For the Android appearance, men (Mdn = 3) reported significantly higher acceptance (U = 108,832.50, $z = -3.94$, $p < .001$) than women (Mdn = 2). This difference remained statistically significant after Bonferroni correction. No significant gender differences were observed for the other appearance types, including Hybrid Android, Robotic High Anthropomorphic, Robotic Low Anthropomorphic, Cartoonish, or Animal-like avatars ($ps > .05$).

These findings suggest that acceptance of robot avatar appearances is meaningfully influenced by cultural identity and, to a lesser extent, by gender.

*3.2.2. Reasons behind android acceptance and rejection*

Given that we were particularly interested in understanding the reasons behind acceptance or rejection of ultra-realistic androids, we included an additional open-ended question to further enquire about participant's choice (agree – neutral – disagree). A thematic analysis was conducted on 663 open-ended responses from participants who either agreed (n = 505) or disagreed (n = 158) with the deployment of ultra-realistic androids in customer service roles. Neutral responses were excluded. Open-ended responses were initially analyzed using

AI-assisted methods to identify recurring semantic patterns and identify key themes by grouping responses based on keyword similarity and latent linguistic features using lexical matching and semantic proximity. We then reviewed and refined the preliminary themes to obtain the final analysis.

Among those who agreed, four dominant themes emerged. The first theme, *Comfort and Familiarity*, reflected the perception that human-like robots made interactions feel more natural and less intimidating. For example, one participant stated, "I feel like it will be more comfortable to talk if its human like" (P24), while another noted, "it makes it more homely" (P149). The second theme, *Alignment with Dubai's Innovation Agenda*, emphasized the congruence between human-like androids and Dubai's identity as a technologically advanced, future-oriented city. One respondent remarked, "Dubai is a cosmopolitan city that embraces innovation and futuristic technologies. Deploying ultra-realistic androids for customer service aligns with this forward-thinking image and can attract tourists and residents" (P15). The third theme, *Enhanced Customer Experience and operational efficiency*, included references to service effectiveness enabled by realistic appearances (e.g., "To have a feeling of more realistic customer support", P230; "They are fast and require less time to solve the issue. Financially beneficial over the humans. They can work 365 days without resting.", P163). Finally, the fourth theme, *Public Appeal and Child Engagement*, highlighted the attractiveness of androids for diverse audiences, including children: "So that people can relate and not be scared of them especially children" (P654).

Among participants who disagreed, four opposing themes were identified. The first theme, *Uncanny Valley and Emotional Discomfort*, expressed unease with overly human-like robots, such as "Human like robots make me feel uncomfortable and a bit scary, I have a feeling that I am talking with a dead person" (P31) and "It's more realistic and less creepy for Robots to look like machines rather than humans" (P40). The second theme, *Job Displacement*, emphasized concerns about automation reducing employment opportunities: "It takes jobs away from actual people who do these jobs now" (P43). The third theme, *Clear Human-Robot Distinction*, reflected the belief that robots should remain visually and functionally distinct from humans, as illustrated by "The robot should appear like a robot not similar to the humans" (P188) and "I will like robot to be distinctive and not to be confused with humans from far or near" (P53). The final theme, *Ethical, Moral, Philosophical, and Religious Concerns*, encompassed deeper ethical, moral, and philosophical reservations about the societal implications of anthropomorphizing machines: "I believe everyone is unique in his own way. I see no reason why the robot should look like a human, it should also have its own unique appearance. Also there will be a thin line between humans and ultra realistic androids that we should keep away from because we then loose the real sense of reality. We should take into consideration the younger generations to come should be able to identify and differentiate human from robots. I know having robots are part of the future everything changes so fast that if they look like us too it will confuse many people." (P384), as well as religious: "God created humans and robots should not resemble us for that reason. And I feel better with a robot that doesn't have human features (P701) and "I have nothing against it, I just think it's not good to imitate humans. God honored humans, created things that resemble them, for me it's denigrating them." (P695).

### 3.3. Spaces in the service sector

To address RQ3, regarding the extent to which robot avatars are accepted in the different spaces of the service sector among Dubai residents, participants were asked to rate their agreement with the deployment of robot avatars across 20 different settings. The results revealed considerable variation in acceptance depending on the type of space. Commercial and retail environments received the highest levels of approval, with 74.5% of respondents agreeing with robot deployment in shopping malls, followed by 61.8% in supermarkets, 60.2% in stores, and 59.6% in hotels. Similarly high acceptance was found in transport and transit hubs, with 69.6% agreement for airports and 68.9% for train and metro stations. Cultural and public leisure spaces also showed favorable attitudes, with 69.1% agreement in museums, 61.9% in libraries, and 58.1% in public parks and streets. In contrast, more moderate acceptance was reported for government and administrative services: 57.9% agreed with robot use in post offices, 49.6% in government offices, and 48.7% in banks. Attitudes were mixed in educational institutions, with 50.4% agreement for language academies, 49.7% for universities, and a notably lower 39.4% for schools and high schools. The lowest levels of acceptance were consistently found in healthcare and wellness settings. Only 37.4% of respondents agreed with robot avatars in hospitals and clinics, 35.9% in rehabilitation centers, 35.3% in pharmacies, 32.3% in nursing homes, and 31.6% in dental clinics. These results suggest that robot avatars are largely accepted in commercial, transport, and cultural spaces, where their roles are likely perceived as task-oriented and non-intrusive. Conversely, the hesitancy observed in government, educational, and especially healthcare settings may reflect concerns about trust, privacy, and the suitability of robotic agents in emotionally sensitive environments. Table 2 shows the full distribution of agreement, neutrality, and disagreement across all service settings included in the study.

| Setting | Agree (%) | Neutral (%) | Disagree (%) |
| --- | --- | --- | --- |
| **Commercial and Retail** | | | |
| Shopping malls | 74.5 | 16.3 | 9.2 |
| Supermarkets | 61.8 | 25.4 | 12.8 |
| Stores | 60.2 | 26.8 | 13.0 |
| Hotels | 59.6 | 26.4 | 14.0 |
| **Transport** | | | |
| Airports | 69.6 | 19.6 | 10.8 |
| Train & metro stations | 68.9 | 20.9 | 10.2 |
| **Cultural and Public Leisure** | | | |
| Museums | 69.1 | 20.2 | 10.7 |
| Libraries | 61.9 | 26.5 | 11.6 |
| Public parks/streets | 58.1 | 27.1 | 14.8 |
| **Government and Administrative Services** | | | |
| Post offices | 57.9 | 28.1 | 14.0 |
| Government offices | 49.6 | 30.6 | 19.8 |
| Banks | 48.7 | 30.1 | 21.2 |
| **Educational Institutions** | | | |
| Language academies | 50.4 | 29.8 | 19.8 |
| Universities | 49.7 | 31.0 | 19.3 |
| Schools & high schools | 39.4 | 33.0 | 27.6 |
| **Event and Professional Venues** | | | |
| Conference centers | 53.5 | 30.2 | 16.3 |
| **Healthcare and Wellness** | | | |
| Hospitals & clinics | 37.4 | 34.3 | 28.3 |
| Rehabilitation centers | 35.9 | 35.4 | 28.7 |
| Pharmacies | 35.3 | 34.5 | 30.2 |
| Nursing homes | 32.3 | 36.5 | 31.2 |
| Dental clinics | 31.6 | 33.1 | 35.3 |

*Table 2. Acceptance of robot avatars across service sector settings (N = 1001)*

### 3.4. Tasks for the service sector

Regarding RQ4, which asked to what extent robot avatars are accepted to perform the different tasks typically performed by social robots, the data revealed a generally favorable attitude among Dubai residents toward robot avatars performing a wide array of tasks commonly associated with social robots. The highest levels of acceptance were observed for utilitarian service-oriented tasks, such as providing information to customers (72.5% agreement), offering guidance to find a place (71.7%), and managing recycling collection (68.5%). Similarly, carrying personal items like shopping bags or suitcases (67.9%) and offering multilingual support (66.7%) were also widely accepted. Tasks related to procedural or administrative support, such as customer registration/check-in (66.0%) and collecting feedback (65.7%), received comparable endorsement.

However, more socially nuanced or emotionally sensitive roles yielded lower levels of acceptance. In particular, only 50.2% agreed that robot avatars could effectively provide companionship and entertainment, and even fewer respondents supported their use for handling customer complaints (43.2%), the lowest among all tasks. Intermediate acceptance was noted for object delivery (58.9%), loyalty program sign-ups (56.7%), security patrolling (55.6%), and telepresence-based service interactions (55.4%).

Overall, the findings suggest that while functional and informational roles for robot avatars enjoy broad support, tasks requiring empathy, discretion, or human rapport evoke more ambivalence or resistance among participants. Table 6 presents the percentage of respondents who agreed, disagreed, or remained neutral regarding the use of robot avatars to perform the different tasks.

| Task | Agree (%) | Neutral (%) | Disagree (%) |
| --- | --- | --- | --- |
| **Provide information to customers** | 72.5 | 18.1 | 9.4 |
| **Provide guidance to find a place** | 71.7 | 22.7 | 5.6 |
| **Recycling collection** | 68.5 | 21.1 | 10.4 |
| **Carry shopping bags or suitcase** | 67.9 | 21.8 | 10.3 |
| **Multilingual support** | 66.7 | 26.9 | 6.4 |
| **Customer registration / check-in – out** | 66 | 23.4 | 10.6 |
| **Customer feedback and surveys** | 65.7 | 24.6 | 9.7 |
| **Object delivery** | 58.9 | 28.1 | 13 |
| **Sign up to loyalty programs** | 56.7 | 30.8 | 12.5 |
| **Patrolling spaces for security** | 55.6 | 27.6 | 16.8 |
| **Telepresence with a service representative** | 55.4 | 27.1 | 17.5 |
| **Companionship and entertainment** | 50.2 | 32.2 | 17.6 |
| **Handle customer complaints** | 43.2 | 28 | 28.9 |

*Table 3.* Acceptance of Robot Avatars for Various Social Robot Tasks among Dubai Residents (N = 1001)

## 4. Discussion

This study examined public acceptance of cybernetic avatars—both physical robots and digital representations—in the multicultural environment of Dubai. The central goal was to understand how acceptance varies across key dimensions, including modality, visual appearance, deployment context, functional role, and demographic background. The research was motivated by the growing integration of cybernetic systems into public services and the need to align technological deployment with societal expectations in diverse communities.

Overall, the results reveal that acceptance of cybernetic avatars in Dubai is relatively high. A more detailed analysis revealed a preference for physical robots over digital avatars. Also, highly anthropomorphic robotic avatars were the most favored appearance type, followed by cartoonish and android designs. Hybrid androids and animal-like robots were less accepted, indicating discomfort with ambiguous or zoomorphic forms. Acceptance also varied significantly by context: commercial, transit, and cultural environments were considered appropriate venues for robot deployment, while schools and healthcare settings received much lower levels of support. These findings suggest that citizens are comfortable with robotic systems for customer service but might be more reluctant to adopt them in spaces that involve more vulnerable populations such as kids, patients, or older adults.

Functional roles also influenced acceptance. Participants expressed strong support for service-oriented and informational tasks, such as guidance and multilingual assistance, but showed greater hesitation toward emotionally sensitive functions like companionship or handling complaints. This pattern reinforces the importance of matching avatar functionality with perceived social appropriateness.

Demographic analyses highlighted that cultural background shapes attitudes toward avatar appearance. Emirati participants showed significantly higher acceptance of android appearances, and the Other Asia cluster showed significantly lower acceptance. For the hybrid android appearance, the Other Asian cluster reported significantly lower acceptance. For cartoonish appearances, the Other Asia cluster were significantly more receptive, and Westerners showed significantly lower acceptance.

Gender analyses revealed that men were significantly more accepting of androids, while no other appearance or modality showed gender-based differences.

These findings carry practical implications for the development and deployment of cybernetic systems. Adaptability in visual style may help align robots with the preferences of specific communities. Moreover, deployment strategies should consider the function and location of robots, ensuring that they are introduced in contexts where they are likely to be perceived as beneficial and appropriate and be particularly mindful when they are intended for interactions in spaces that involve vulnerable populations.

Several limitations should be acknowledged. First, the study relied on still images to represent avatar appearances, which may not fully capture the dynamic and behavioral qualities that influence real-world acceptance. Second, the survey context was hypothetical, asking participants to imagine future scenarios rather than responding to direct interaction. In the future, real exposure to robots may lead to different results. In this regard, it would be particularly valuable to conduct a longitudinal follow-up to examine how public attitudes evolve over time as individuals gain firsthand experience with these systems in real-life settings.

Our study offers a unique contribution to the field by capturing perspectives from residents of over 80 countries. To our knowledge, no previous work in the domain of human–robot interaction has addressed the topic of avatar acceptance with a similar breadth of cultural representation. The multicultural structure of Dubai allowed us to gather views from highly diverse backgrounds within a single urban setting, enabling a comparative lens that is rarely accessible in conventional population samples. This cross-cultural richness positions the present dataset as an empirical foundation for future research seeking to generalize findings across global contexts.

As society increasingly consider integrating robotic and virtual agents into service infrastructures, understanding public sentiment is essential. By identifying where and how different formats and appearances are accepted—or resisted—this research contributes actionable insights for inclusive and culturally sensitive deployment. Importantly, the implications of this research extend beyond survey data collection. This study has been supported also as part of a concrete strategy to guide real-world deployments. The insights gained are intended to directly inform the future design and implementation of cybernetic avatars—both robotic and digital—across public and private service sectors in the country. In this sense, the project exemplifies science-driven innovation: rigorous empirical research serving as the foundation for immediate societal impact.

While this study centers on cybernetic avatars, the findings are also relevant to the domains of social and humanoid robotics, which might trigger similar acceptance as robot avatars, particularly concerning appearance, settings, and tasks. As such, future research could also expand on our findings to refine the conceptual boundaries of the "uncanny valley" in multicultural contexts.

As robots become increasingly integrated into various aspects of daily life, understanding public acceptance of cybernetic avatars and social robots in general is crucial for a successful deployment. Incorporating citizen feedback into the development and deployment of these technologies throughout all stages is critical to ensure a harmonious coexistence between humans and robots in future's society.

### 5. Declaration of generative AI in the writing process

During the preparation of this work the author(s) used ChatGPT to improve readability of the manuscript. After using this tool/service, the author(s) reviewed and edited the content as needed and take(s) full responsibility for the content of the publication.

**SUPPLEMENTARY INFORMATION**

**Table 1A**

*Binomial Test Results for Physical Robot Acceptance by Community Cluster*

| Cluster | Agree (%) | Test Proportion | Exact Sig. (1-tailed) |
| --- | --- | --- | --- |
| Emirati | 74.1 | 0.673 | .033 |
| Middle East | 70.5 | 0.673 | .216 |
| South Asia | 67.1 | 0.673 | .501 |
| Other Asia | 63.8 | 0.673 | .182 |
| Western | 63.4 | 0.673 | .163 |
| Other Africa | 65 | 0.673 | .293 |

**Table 2A**

*Binomial Test Results for Digital Avatar Acceptance by Community Cluster*

| Cluster | Agree (%) | Test Proportion | Exact Sig. (1-tailed) |
| --- | --- | --- | --- |
| Emirati | 64.7 | 0.569 | .023 |
| Middle East | 51.2 | 0.569 | .081 |
| South Asia | 48 | 0.569 | .011 |
| Other Asia | 55.2 | 0.569 | .350 |
| Western | 60.9 | 0.569 | .174 |
| Other Africa | 62.4 | 0.569 | .094 |

**Table 3A**

*Binomial Test Results for Android Appearance Agreement by Community Cluster*

| Cluster | Agree (%) | Test Proportion | Exact Sig. (1-tailed) |
|---|---|---|---|
| Emirati | 69.4 | 0.504 | < .001 |
| Middle East | 53.6 | 0.504 | .226 |
| South Asia | 46.2 | 0.504 | .154 |
| Other Asia | 33.3 | 0.504 | < .001 |
| Western | 50.3 | 0.504 | .522 |
| Other Africa | 50.3 | 0.504 | .524 |

*Note.* "Agree" responses were coded as successes in the binomial test.

**Table 4A**

*Binomial Test Results for Hybrid Android Appearance Agreement by Community Cluster*

| Cluster | Agree (%) | Test Proportion | Exact Sig. (1-tailed) |
|---|---|---|---|
| Emirati | 45.3 | 0.412 | .157 |
| Middle East | 44.6 | 0.412 | .210 |
| South Asia | 41.0 | 0.412 | .516 |
| Other Asia | 31.6 | 0.412 | .006 |
| Western | 40.4 | 0.412 | .449 |
| Other Africa | 44.6 | 0.412 | .217 |

*Note.* "Agree" responses were coded as successes in the binomial test.

**Table 5A**

*Binomial Test Results for Cartoonish Appearance Agreement by Community Cluster*

| Cluster | Agree (%) | Test Proportion | Exact Sig. (1-tailed) |
|---|---|---|---|
| Emirati | 57.6 | 0.533 | .145 |
| Middle East | 50.0 | 0.533 | .219 |
| South Asia | 48.0 | 0.533 | .092 |
| Other Asia | 68.4 | 0.533 | < .001 |
| Western | 40.4 | 0.533 | < .001 |
| Other Africa | 54.8 | 0.533 | .386 |

*Note.* "Agree" responses were coded as successes in the binomial test.

**Table 6A**

*Binomial Test Results for Robotic-Looking, Highly Anthropomorphic Appearance Agreement by Community Cluster*

| Cluster | Agree (%) | Test Proportion | Exact Sig. (1-tailed) |
|---|---|---|---|
| Emirati | 67.1 | 0.611 | .064 |
| Middle East | 58.4 | 0.611 | .265 |
| South Asia | 56.1 | 0.611 | .101 |
| Other Asia | 61.5 | 0.611 | .491 |
| Western | 65.8 | 0.611 | .124 |
| Other Africa | 58.0 | 0.611 | .233 |

*Note.* "Agree" responses were coded as successes in the binomial test.

**Table 7A**

*Binomial Test Results for Robotic-Looking, Low Anthropomorphic Appearance Agreement by Community Cluster*

| Cluster | Agree (%) | Test Proportion | Exact Sig. (1-tailed) |
|---|---|---|---|
| Emirati | 50.0 | 0.426 | .031 |
| Middle East | 44.0 | 0.426 | .388 |
| South Asia | 34.7 | 0.426 | .020 |
| Other Asia | 48.9 | 0.426 | .056 |
| Western | 37.9 | 0.426 | .129 |
| Other Africa | 39.5 | 0.426 | .240 |

*Note.* "Agree" responses were coded as successes in the binomial test.

**Table 8A**

*Binomial Test Results for Animal-Like Appearance Agreement by Community Cluster*

| Cluster | Agree (%) | Test Proportion | Exact Sig. (1-tailed) |
|---|---|---|---|
| Emirati | 46.5 | 0.390 | .028 |
| Middle East | 31.9 | 0.390 | .036 |
| South Asia | 38.2 | 0.390 | .442 |
| Other Asia | 44.3 | 0.390 | .090 |
| Western | 31.7 | 0.390 | .033 |
| Other Africa | 40.8 | 0.390 | .353 |

*Note.* "Agree" responses were coded as successes in the binomial test.

**Table 9A**

*Gender Differences in Preferences for Robot Avatar Appearance*

| Appearance Type | Z | p |
|---|---|---|
| Android | -3.94 | .001 |
| Hybrid Android | -.99 | .319 |
| Robotic-Looking, High Anthro | -.15 | .877 |
| Robotic-Looking, Low Anthro | -.50 | .615 |
| Cartoonish | -1.27 | .202 |
| Animal-Looking | -.60 | .548 |